\documentclass[10pt,twocolumn,letterpaper]{article}

\usepackage[pagenumbers]{cvpr} % To force page numbers, e.g. for an arXiv version

\usepackage{graphicx}
\usepackage{amsmath}
\usepackage{amssymb}
\usepackage{booktabs}
\usepackage[ruled]{algorithm2e}
\usepackage{multirow}
\usepackage{times}
\usepackage{epsfig}

\usepackage[pagebackref,breaklinks,colorlinks]{hyperref}

\usepackage[capitalize]{cleveref}
\crefname{section}{Sec.}{Secs.}
\Crefname{section}{Section}{Sections}
\Crefname{table}{Table}{Tables}
\crefname{table}{Tab.}{Tabs.}

\def\cvprPaperID{8451} % *** Enter the CVPR Paper ID here
\def\confName{CVPR}
\def\confYear{2023}

\begin{document}

\title{Feature Extraction Matters More: Universal Deepfake Disruption through Attacking Ensemble Feature Extractors}

%\author{
%    \IEEEauthorblockN{San Zhang$^{a*}$, Si Li$^{a,b}$, Wu Wang$^b$}
%    \IEEEauthorblockA{$^a$ School of Computer Science, Wuhan University, Wuhan, China}
%    \IEEEauthorblockA{$^b$ Department of Computer Science and Technology, Tsinghua University, Beijing, China}
%    \IEEEauthorblockA{\{zhangsan\}@XXX.com, \{lisi, wangwu\}@XXX.edu.cn}
%}

\author{Long Tang$^{1,2}$,Dengpan Ye$^{1,2*}$,Zhenhao Lu$^{1,2}$,Yunming Zhang$^{1,2}$,Shengshan Hu$^{3}$,Yue Xu$^{1,2}$,Chuanxi Chen$^{1,2}$\\
$^{1}$School of Cyber Science and Engineering,Wuhan University,China\\
$^{2}$Key Laboratory of Aerospace Information Security and Trusted Computing, Ministry of Education, China\\
$^{3}$School of Cyber Science and Engineering, Huazhong University of Science and Technology Wuhan, China\\
{\tt\small \{l\_tang,yedp\}@whu.edu.cn}
}
\maketitle

%%%%%%%%% ABSTRACT
\begin{abstract}
Adversarial example is a rising way of protecting facial privacy security from deepfake modification. To prevent massive facial images from being illegally modified by various deepfake models, it is essential to design a universal deepfake disruptor. However, existing works treat deepfake disruption as an End-to-End process, ignoring the functional difference between feature extraction and image reconstruction, which makes it difficult to generate a cross-model universal disruptor. In this work, we propose a novel \textbf{F}eature-\textbf{O}utput ensemble \textbf{UN}iversal \textbf{D}isruptor (FOUND) against deepfake networks, which explores a new opinion that considers attacking feature extractors as the more critical and general task in deepfake disruption. We conduct an effective two-stage disruption process. We first disrupt multi-model feature extractors through multi-feature aggregation and individual-feature maintenance, and then develop a gradient-ensemble algorithm to enhance the disruption effect by simplifying the complex optimization problem of disrupting multiple End-to-End models. Extensive experiments demonstrate that FOUND can significantly boost the disruption effect against ensemble deepfake benchmark models. Besides, our method can fast obtain a cross-attribute, cross-image, and cross-model universal deepfake disruptor with only a few training images, surpassing state-of-the-art universal disruptors in both success rate and efficiency.
\end{abstract}

%%%%%%%%% BODY TEXT
\section{Introduction}\label{sec1}

Since the Generative Adversarial Network(GAN) was proposed, it has demonstrated huge potential in perceiving, analyzing, and reconstructing multimedia data. It has made significant development in virtual content generation, image processing, etc., creating considerable social benefits. However, some criminals use GAN to tamper users' private images into a target person or false content, resulting in increasingly widespread attention to the privacy and reputation risks caused by GAN-based image tampering (also called DeepFakes)\cite{juefei2022countering}. It is necessary to develop effective strategies against deepfake tampering.

The methods of mitigating deepfake risks mainly include passive defense and active defense. Passive deepfake detection is a kind of post-defense method\cite{wang2020cnn,wang2021fakespotter,hu2021exposing,bonettini2021video,tariq2021one}. It judges whether it is a fake image by training a binary forgery detector. However, this method cannot completely prevent the impact and harm of forged images. Active defense\cite{ruiz2020disrupting,huang2021initiative,wang2022anti,huang2022cmua,wang2022deepfake} destroys the deepfake generator by superimposing antagonistic perturbations that cannot be recognized by the human eye on the protected image, so as to produce an obviously unreal output. This method can protect the image from tampering fundamentally. In order to protect more images from being tampered by different deepfake methods, it is easy to consider generating a universal deepfake disruptor. To obtain a universal deepfake disruptor, existing works require sequential attacks against all possible attribute modifications of an image\cite{ruiz2020disrupting} and require a large number of queries to select appropriate weights to fuse perturbations of multiple models\cite{huang2022cmua}. On the other hand, most of the existing research regards attacking deepfake models as an End-to-End process. However, deepfake models contain the feature extraction module and the generation module, different deepfake models modify feature attributes in different stages of the generation process. The complexity and difference of both the structure and attribute modification lead to a huge semantic gap between deepfake models, making it hard to attack directly.

Based on the above inferences, we rethink the process of deepfake disruption. We consider the feature extraction module of the deepfake model the key to determining the quality of image generation, then propose a novel \textbf{F}eature-\textbf{O}utput ensemble \textbf{UN}iversal \textbf{D}isruptor (FOUND) against deepfake models. First, we conduct an ensemble attack on the feature extractors of multiple models to destroy their feature extraction capabilities and then conduct an End-to-End ensemble attack on the entire models. We just attack the original attributes of the image without traversing all target attributes. Finally, we decompose the complex optimization problem into multiple simple tasks, and propose a novel gradient-ensemble attack to alleviate optimization conflict between End-to-End ensemble models.

Our contributions are summarized as follows:
\begin{itemize}
    \item[$\bullet$] We are the first to propose attacking the feature extraction module in deepfake disruption. We propose a complete universal attack pattern against multi-deepfake models, and introduce a simple yet effective two-stage deepfake disruption method based on just a few images with their original attributes.
    \item[$\bullet$] Different from existing works, we propose a novel gradient-ensemble strategy in the End-to-End ensemble attack to solve the complex multi-deepfake model optimization problem. It prevents the ensemble attack from falling into the local optimum of the easy-to-attack models, and improves the overall effect of the multi-model ensemble disruption.
    \item[$\bullet$] Experimental results show that attacking feature extractors can significantly improve the disruption effect against ensemble deepfake models, achieving a competitive success rate and significantly higher efficiency than state-of-the-art universal disruptors.
\end{itemize}
%-------------------------------------------------------------------------

% Update the cvpr.cls to do the following automatically.
% For this citation style, keep multiple citations in numerical (not
% chronological) order, so prefer \cite{Alpher03,Alpher02,Authors14} to
% \cite{Alpher02,Alpher03,Authors14}.

%------------------------------------------------------------------------
\section{Related Works}\label{sec2}
\subsection{Deepfake Methods}
DeepFakes can be generally divided into four types: entire synthesis\cite{Karras_2019_CVPR,NEURIPS2021_076ccd93}, attribute modification\cite{choi2018stargan,he2019attgan,tang2019attention,li2021image}, identity swap\cite{Korshunova_2017_ICCV,peng2021unified}, and face reenactment\cite{Garrido_2014_CVPR,Thies_2016_CVPR}. We focus on attribute modification methods. They use the GAN network to modify facial attributes to achieve refined facial manipulations. StarGAN\cite{choi2018stargan} proposes a scalable method to perform image conversion across multiple attributes through one GAN backbone. Subsequently, the authors further propose StarGAN-v2 \cite{choi2020stargan}, which can extract the style code of the target image and modify a variety of styles, so as to obtain more diverse attribute editing effects. AttGAN\cite{he2019attgan} uses attribute classification constraints to provide a more natural facial image of facial attribute operations. AGGAN\cite{tang2019attention} uses the built-in attention mechanism to introduce an attention mask to constrain the attribute editing area, so as to obtain high-quality forged images. HiSD\cite{li2021image} is an advanced image-to-image attribute editing method, which can extract multiple specific attributes on the target image and embed them on the original image. These models vary in model structure, loss, and attribute editing tasks. It is difficult to disrupt them simultaneously.

%-------------------------------------------------------------------------
\subsection{Deepfake Disruption}
Several studies have been done on exploring adversarial attacks on generative models\cite{ruiz2020disrupting,huang2021initiative,wang2022anti,huang2022cmua,wang2022deepfake}. Ruiz \textit{et al.}\cite{ruiz2020disrupting} is the first to disrupt conditional image translation networks through adversarial examples. Huang \textit{et al.}\cite{huang2021initiative} propose a disruption framework by training a surrogate model to attack in the grey-box settings. Wang \textit{et al.}\cite{wang2022anti} propose a robust deepfake disruption attack by converting the attack process from RGB to Lab color space. All these generated perturbation only works on a single image. CMUA-Watermark\cite{huang2022cmua} is the first work proposing a universal adversarial perturbation to deepfake disruption, but it relies on third-party tools and requires a lot of iterations, which is time-consuming. Our proposed method performs parallel attacks and does not need to perform attacks on all target attributes of each image, obtaining better universality and efficiency.

\begin{figure}[t]
    \centering
\epstopdfsetup{outdir=./}
    \includegraphics[width=0.8\linewidth]{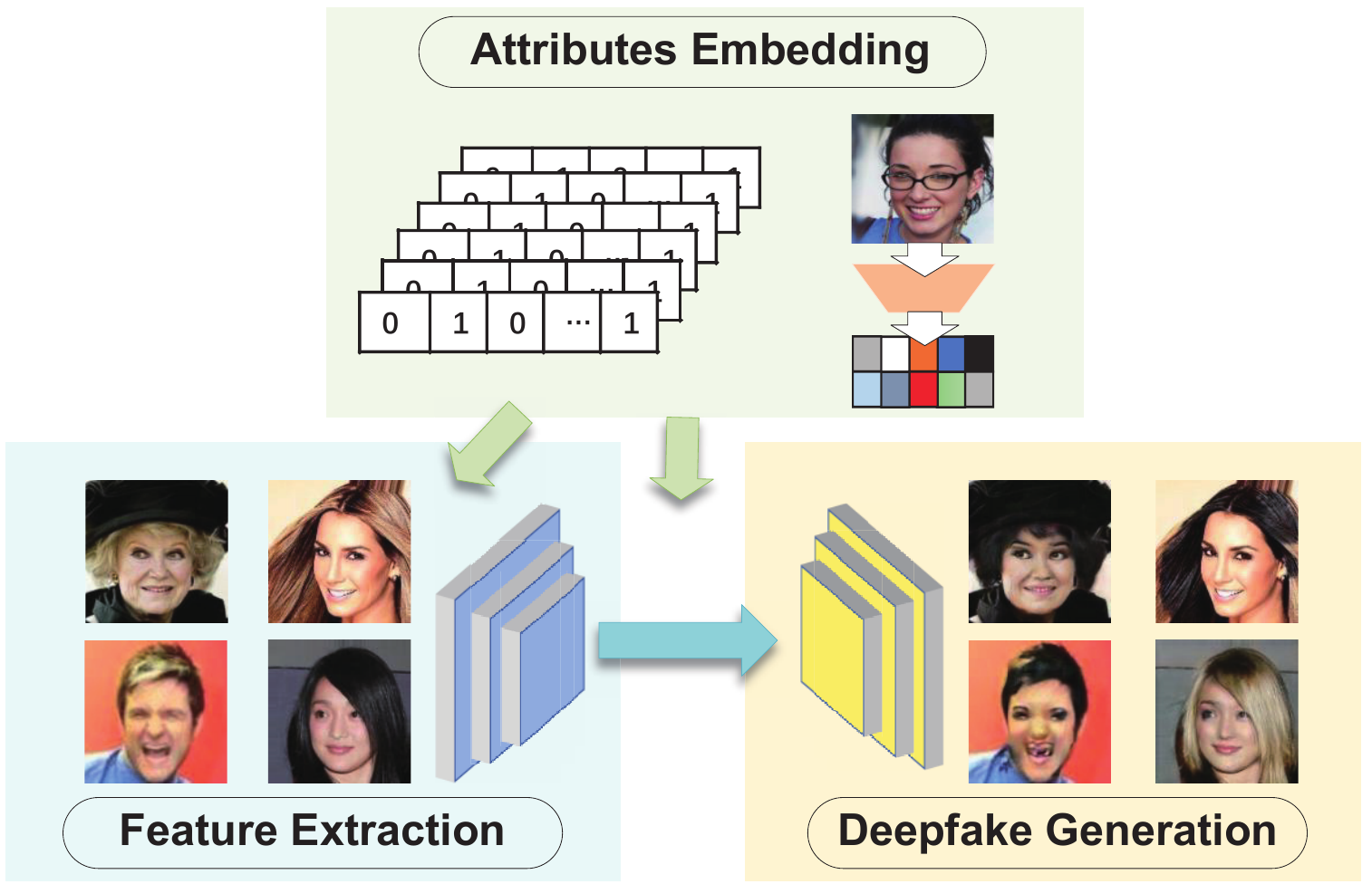}
    \caption{The typical process of attribute modification networks.}
\label{fig1}
\end{figure}

%-------------------------------------------------------------------------
\subsection{Universal Adversarial Perturbation}
Moosavi Dezfouli \textit{et al.}\cite{moosavi2017universal} first propose the concept of universal adversarial perturbation, by which the results of a classification model on multiple images can be fooled with only one adversarial perturbation. Based on this work, Metzen \textit{et al.}\cite{hendrik2017universal} and Li \textit{et al.}\cite{li2019universal} introduce generic adversarial perturbations in segmentation tasks and image retrieval tasks, respectively. These works only produce a universal adversarial example targeting one model. Che \textit{et al.}\cite{che2020smgea} propose a cross-task universal attack method SMGEA against multiple models with different tasks by fusing the intermediate features and using the long- and short-term memory gradients, but the generated adversarial example cannot work across different images. 

%-------------------------------------------------------------------------
% \begin{table}
%   \centering
%   \begin{tabular}{@{}lc@{}}
%     \toprule
%     Method & Frobnability \\
%     \midrule
%     Theirs & Frumpy \\
%     Yours & Frobbly \\
%     Ours & Makes one's heart Frob\\
%     \bottomrule
%   \end{tabular}
%   \caption{Results.   Ours is better.}
%   \label{tab:example}
% \end{table}
%-------------------------------------------------------------------------

\section{Motivation}\label{sec3}
As shown in Figure \ref{fig1}, a typical attribute modification network mainly includes feature extraction, target attribute embedding, and deepfake generation. The difference between deepfake models lies in the content and insert stage of the embedded target attributes. In this section, we use StarGAN\cite{choi2018stargan}, AttGAN\cite{he2019attgan} and HiSD\cite{li2021image} as an example to analyze the attribute modification networks, and explain why it is unnecessary to use all target attributes to attack deepfake models.

%-------------------------------------------------------------------------

\begin{figure}[t]
    \centering
	\epstopdfsetup{outdir=./}    
	\includegraphics[width=0.9\linewidth]{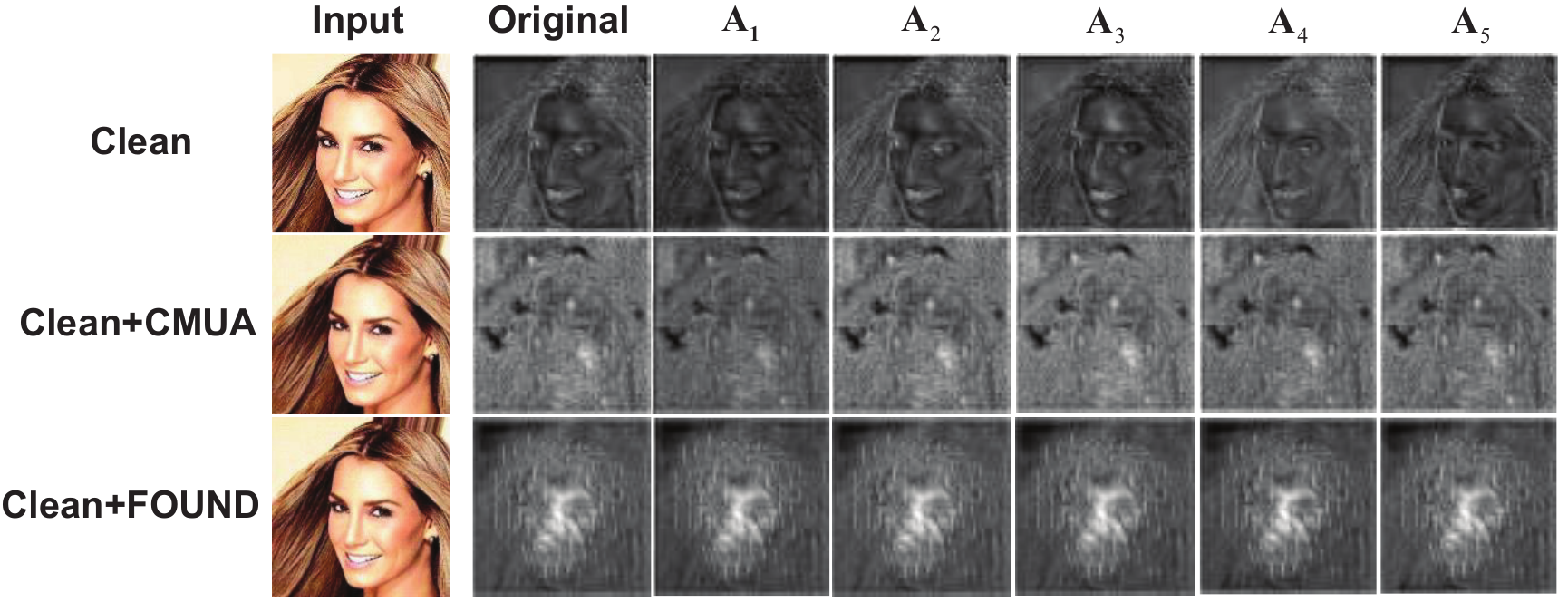}
    \caption{The outputs of feature extractors of StarGAN with different inputs. $Original$ refers to the output features with the ground-truth attribute label, $A_1{\sim}A_5$ refers to the output features with different target attributes.}
\label{fig2.1}
\end{figure}

\subsection{Analysis of Model Heterogeneity}
StarGAN is trained to be able to modify the face attributes through the binary attribute labels. Specifically, StarGAN expands each binary attribute into a feature map of the image size and concatenates with the input image in the channel dimension. That is, the target attributes run through the entire process of deepfake generation. The generation process of StarGAN can be briefly expressed as:
$$\bar{x}=G(E(x,a))$$
where $x$ is the original image, $a$ is the target attribute, $\bar{x}$ is the generated deepfake image, $E(\cdot)$ and $G(\cdot)$ are the feature extractor and generator of the model respectively, the same hereinafter.

AttGAN also modifies face attributes through binary attribute labels, but it concatenates binary attribute feature maps in the generator’s first layer, which means the attribute modification only influent the generation process. The generation process of AttGAN can be briefly expressed as:
$$\bar{x}=G(E(x),a)$$

HiSD is different from the above two models. It is an image-to-image translation model which is able to extract a specific attribute from one image and embed it to the other. The target attributes are no longer binary labels. The generation process of HiSD can be briefly expressed as:
$$\bar{x}=G(T(E(x),S(y_a)))$$
where $S(\cdot)$ is the specific attribute extractor, $y_a$ is the image with the target attribute, $T(\cdot)$ is the feature mixup model. The generated $\bar{x}$ is the original image $x$ embedded with tampering target $S(y_a)$.

According to the above descriptions, various deepfake models have different forgery strategies and mechanisms. If we directly treat the deepfake methods as End-to-End models, the significant difference in structures, tasks, and attribute distributions between deepfake models will make it a more complex optimization problem to disrupt multi-DeepFakes than disrupting multi-classifiers.

\begin{figure}[t]
    \centering
\epstopdfsetup{outdir=./}    
\includegraphics[width=0.8\linewidth]{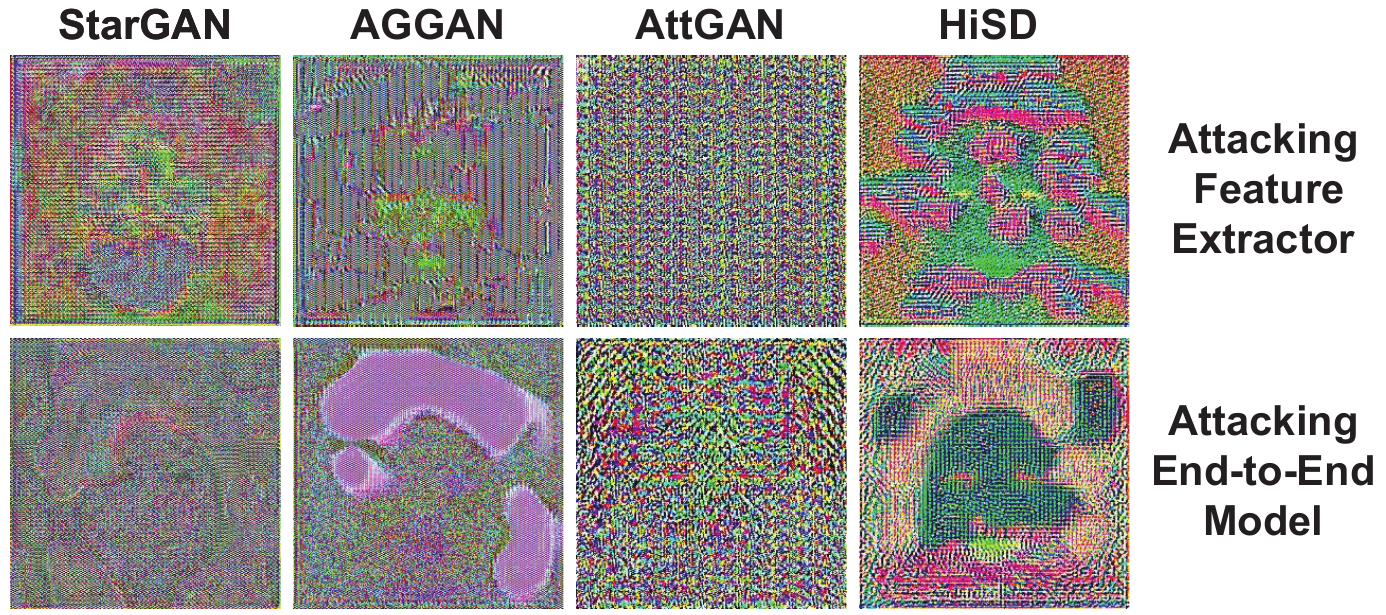}
    \caption{Illustration of adversarial perturbations generated by different models and stages.}
    \label{fig2.2}
\end{figure}

%-------------------------------------------------------------------------
\subsection{Feature Extraction Matters More}

From another perspective, although the attribute embedding process of each model is different, the feature extraction process is similar and is minimally affected by attributes. The first row of Figure \ref{fig2.1} shows the output results of StarGAN's feature extractor under different attribute labels. It can be seen that different attributes have little influence on the feature extraction process. According to the above formulas, AttGAN and HiSD do not involve attribute editing in the feature extraction process. If the model cannot effectively extract the original features, the subsequent generation process cannot get the correct original image information, no matter what kind of target attributes are embedded, high-quality fake images cannot be obtained. The second and third rows in Figure \ref{fig2.1} show the output of the feature extractor of StarGAN with the disruption by CMUA\cite{huang2022cmua} and the proposed FOUND. Compared with CMUA, FOUND causes the feature extractor completely unable to obtain facial information. The experiments in Section \ref{section5} show that FOUND can cause greater disruption in visualization.

Figure \ref{fig2.2} shows the adversarial perturbations generated by different deepfake models and stages respectively. The perturbations of attacking different models vary, but the adversarial perturbations for the feature extractor and the End-to-End process are similar. In Section \ref{section5.3}, we show that just attacking the feature extractor can approach the success rate of attacking the End-to-End model.

\begin{figure*}[htbp]
    \centering
\epstopdfsetup{outdir=./}
    \includegraphics[width=0.9\linewidth]{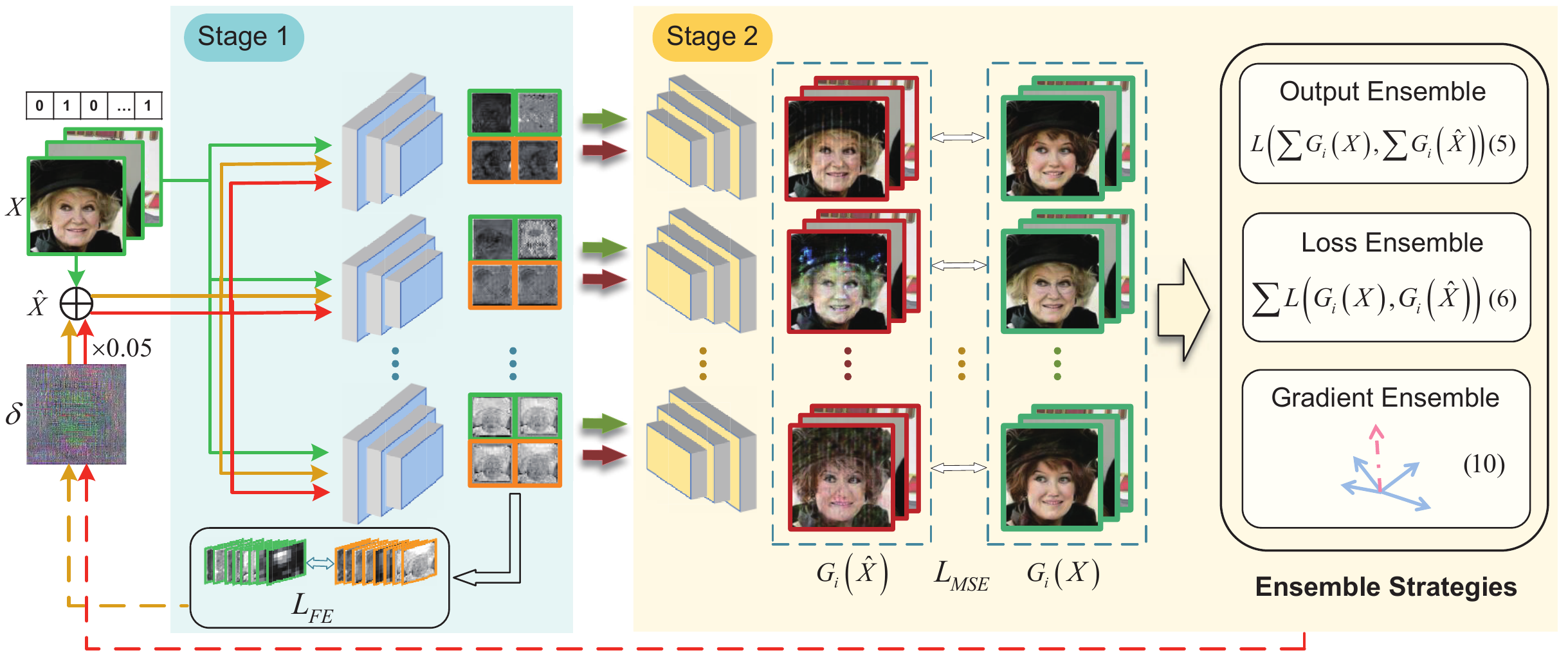}
    \caption{The overall pipeline of FOUND. Green arrows and frames illustrate original images, features, and deepfake images without disruption. Yellow arrows and frames illustrate the process of Feature-Ensemble disruption. Red arrows and frames illustrate the process of End-to-End-Ensemble disruption. The dotted lines illustrate the backward propagation.}
\label{fig3}
\end{figure*}

Therefore, we believe that attacking feature extraction modules matters more in deepfake disruption. On one hand, decomposing the complex problem into small problems can reduce the difficulty of solving, on the other hand, the above theories and phenomena show that disrupting feature extractors will fundamentally destroy the generation quality, making it easier to obtain a universal deepfake disruptor.
%-------------------------------------------------------------------------

\section{Methodology}\label{section4}

Based on the above discussions, we introduce our proposed FOUND. FOUND is a universal perturbation that can protect massive facial images against various deepfake models. The overview of FOUND is illustrated in Figure \ref{fig3}. We directly train and update the deepfake disruptor on an image-size matrix for different images and models, and the disruptor is obtained through a two-stage attack. In each round of iteration, we first execute Stage 1 to attack the extractors of different deepfake models. We fuse the resized features extracted by each model and execute the attack, so as to get the first-level perturbation that can disrupt the feature extractors; Based on this perturbation, we then execute Stage 2 to perform an End-to-End ensemble attack on the whole process of multiple deepfake models. We perform gradient-ensemble with the gradients obtained from different models and use the fused gradient to update the deepfake disruptor.
%-------------------------------------------------------------------------

\subsection{Feature-Ensemble Disruption and Anti-Deviation}\label{sec4.1}
In this section, we introduce how to disrupt the feature extractors of ensemble models(Stage 1 in Figure \ref{fig3}) in detail. First, we initialize the image-size disruptor $\delta$ with Gaussian noise, then add it to a batch of input images $\textit{X}$ to get the adversarial examples $\hat{X}=X+\delta$. 
% \begin{equation}
%     \begin{split}
%         \label{eq0}
%         \hat{X}=X+\delta
%     \end{split}
% \end{equation}
We send the image pair $(X,\hat{X})$ into a set of deepfake models and obtain the output of the feature extractors, respectively. As shown in Section \ref{sec3}, we denote the feature extractors as $E_0(\cdot),E_1(\cdot),...,E_i(\cdot)$. 

After obtaining the extracted features, we need to mix up these multi-model features. Since the feature map resolution and the number of channels obtained by the various feature extractors are different, it is necessary to resize the feature maps and ensure that the operation of the feature space aggregation is differentiable. Therefore, we use differentiable bilinear interpolation to resize feature maps to the same size and then sum up the differentiable elements on all of the feature maps in the channel dimension to obtain the integrated features.

Then, we utilize Mean Square Error(MSE) to measure the differences between the integrated original features and adversarial features,
\begin{equation}
    \begin{split}
        \label{eq1}
        L_{MSE}=MSE(&\sum_{i=1}^{N}BI(E_{i}(X)),\sum_{i=1}^{N}BI(E_{i}(\hat{X}))),\\ &s.t.||\hat{X}-X||_\infty<\epsilon
    \end{split}
\end{equation}
where $\epsilon$ is the perturbation limit, aiming not to influence the visualization of the original images. $BI(\cdot)$ is the bilinear interpolation process, and $N$ is the number of the ensemble feature extractors. We maximize the MSE distance between the original features and adversarial features, making feature extractors fail to extract facial attributes correctly.

Notice that this process is different from SMGEA\cite{che2020smgea}. They select specific intermediate layers and only choose part of the feature maps or some important values in each fusion model for stacking. They use the softmax function to normalize the value to $(0,1)$ for each selected feature. We wish the disruptor have the ability to widely adapt to multiple images, only measuring part of the feature maps cannot completely destroy the feature extractor's encoding ability to massive images. In addition, previous studies in attacking classification models\cite{dong2018boosting} have shown that the aggregation of logits can better capture the gradients of different models, and the softmax operation will smooth the distribution between the data, which will lead to the gradient vanishing problem\cite{sun2020gradient}. Therefore, we directly sum the feature maps up without normalization.

Further, we use Kullback-Leibler(KL) divergence\cite{kullback1951information} to measure the difference between the distributions of feature maps of different models. The feature maps of each model are summed in the channal dimension, and then normalized to $(0,1)$ using softmax function.
\begin{equation}
    \begin{split}
        \label{eq2}
        L_{KL}=\sum_{i=1}^{N}\sum_{j{\neq}i}^{N}KL(M(\sum_{k=1}^{C_i}E_i(\hat{X})),M(\sum_{k=1}^{C_j}E_j(\hat{X})))
    \end{split}
\end{equation}
where $KL(\cdot)$ is the KL divergence function, $M(\cdot)$ is the softmax function, $N$ is the number of models and $C_i$ is the number of channels for the $i_{th}$ model. According to Eq.(\ref{eq2}), we minimize $L_{KL}$ to avoid over-fitting some easy-to-attack models in the process of feature-ensemble disruption, making as much overlap as possible in the probability distribution of intermediate features of each disrupted model, so as to better allow the disruptor to cover the general disruption space, thereby improving the universality of the disruptor.

Hence, the final loss of Feature-Ensemble disruption is,
\begin{equation}
    \centering
    \label{eq3}
    L_{FE}=L_{MSE} - \lambda L_{KL}
\end{equation}
where $\lambda$ is the hyperparameter to adjust the proportion of the loss. After computing the loss according to Eq.(\ref{eq3}), we use the PGD attack\cite{madry2018towards} to iteratively update the disruptor,
\begin{equation}
    \begin{split}
        \label{eq4}
        &\delta_{adv}^{0}=\delta,\\
        \delta_{adv}^{r+1}=clip_{\delta,\epsilon}&(\delta_{adv}^{r}+\alpha sign(\nabla_{\delta}L_{FE}))
    \end{split}
\end{equation}
where $r$ refers to each iteration, $\alpha$ is a hyperparameter referring to the learning rate. Then we finish our first stage attack and obtain the feature-level universal disruptor.
%-------------------------------------------------------------------------

\subsection{Output Disruption with Gradient-Ensemble}\label{sec4.2}
Although feature extractors play a major role in the deepfake model, the generation module also determines the quality of deepfake generation. Based on the disruptor obtained by the feature-level attack, we continue executing Stage 2 to perform End-to-End attacks on the ensemble models to further improve the universal disruption ability. 

Ensemble attacks on classification models have been widely discussed. Common ensemble methods include output ensemble and loss ensemble.

\textbf{Output Ensemble} average the output results of different models first and then compute the loss and gradient sequentially. Since classification models output the predictions of images, Liu \textit{et al.}\cite{liu2016delving} and Dong \textit{et al.}\cite{dong2018boosting} propose to ensemble the predictions(after-softmax) and logits(before-softmax) respectively. In deepfake disruption, the output is a fake image, we do not differentiate prediction-ensemble and logit-ensemble. We compute the Output Ensemble by:
\begin{equation}
    \begin{split}
        \label{eq5}
        G_{OE}=\nabla_{\delta}L(\sum_{i=1}^{N}\omega_i G_i(E_i(X)),\sum_{i=1}^{N}\omega_i G_i(E_i(\hat{X})))
    \end{split}
\end{equation}
where $G_i(\cdot)$ is the output result of the $i$th deepfake model, $w_i\geq 0$ is the same ensemble weight for each model. $\sum_{i=1}^{N}\omega_i=1$, the same hereinafter.

\textbf{Loss Ensemble} average the loss of different models, then compute the gradient using the aggregated loss:
\begin{equation}
    \begin{split}
        \label{eq6}
        G_{LE}=\nabla_{\delta}\sum_{i=1}^{N}\omega_{i}L_i(G_i(E_i(X)),G_i(E_i(\hat{X})))
    \end{split}
\end{equation}
where $L_i(\cdot)$ is the loss computed on model $G_i(\cdot)$.

Li \textit{et al.}\cite{li2021exploring} propose using Output Ensemble and Loss Ensemble to attack StyleGAN generator and forensic networks and have achieved a competitive performance. But experiments show that when the gradients between the ensemble models are significantly different, it is very difficult to use these two ensemble methods to compute simultaneously valid gradients. 

We further formalize the problem. Let $F_i(x)$ be the $i$th model with input $x$, we compute the gradient of $N$ ensemble models with Output Ensemble by:
\begin{equation}
    \begin{split}
        \label{eq7}
        g=\frac{\partial L(\sum_{i=1}^{N}F_i(x))}{\partial x}
    \end{split}
\end{equation}
and we perform Loss Ensemble by:
\begin{equation}
    \begin{split}
        \label{eq8}
        g=\frac{\partial \sum_{i=1}^{N}L_i(F_i(x))}{\partial x}
    \end{split}
\end{equation}
With the increasing of $N$ and the larger structure gaps between $F_i(\cdot)$, the composite function becomes so complex that it is hard to compute the gradient and is easy to fall into the local optimum.

Inspired by meta-learning\cite{hospedales2021meta}, we propose the \textbf{Gradient Ensemble} strategy. We separate the entire optimization task into mini-tasks, computing the gradient on each mini-task and fuse them to get the ensemble gradient,
\begin{equation}
    \begin{split}
        \label{eq9}
        g=\sum_{i=1}^{N}\frac{\partial L_i(F_i(x))}{\partial x}
    \end{split}
\end{equation}
In this way, we avoid the gradient falling into the local optimum of some easy-to-optimize functions, and the aggregated gradients can eliminate the preference for a certain model, so as to obtain a better ensemble gradient direction.

\begin{algorithm}[htpb]
	\caption{Feature-Output ensemble UNiversal Disruptor(FOUND)}
	\label{ag1}
	\LinesNumbered 
	\KwIn{$X$(training facial images), $G_i(E_i(\cdot))$(deepfake models $G$ and feature extractors $E$), $T$(iteration number of End-to-End attack), $K$(iteration number of Feature-Ensemble attack)}
	\KwOut{Feature-Output ensemble Universal Disruptor $\delta$}
	Random-Init $\delta$\;
	\For{$t$ in T}{
        \For{$x_b$ in {X}}{
            \For{$k$ in K}{
                Compute Feature-Ensemble loss on $E_i(x_b+\delta)$ with Eq.(\ref{eq3})\;
                Update $\delta$ with Eq.(\ref{eq4})\;
            }
            Compute Gradient-Ensemble on $G_i(x_b+\delta)$ with Eq.(\ref{eq10})\;
            Update $\delta$ with gradient-based adversarial attacks\;
        }
	}
	\Return $\delta$
\end{algorithm}

Finally, we consider disrupting each model as a mini-task and rewrite Eq.(\ref{eq9}) to Eq.(\ref{eq10}) to perform our proposed End-to-End Gradient Ensemble disruption.
\begin{equation}
    \begin{split}
        \label{eq10}
        G_{GE}=\sum_{i=1}^{N}\frac{1}{N}\nabla_{\delta} L_i(G_i(E_i(X)),G_i(E_i(\hat{X})))
    \end{split}
\end{equation}

In Stage 2, we continue using MSE loss to measure and maximize the distance between original deepfake images and disrupted deepfake images. After obtaining the ensemble gradient $G$, we use gradient-based adversarial attacks to iteratively update the disruptor. Most of the popular gradient-based adversarial attacks\cite{kurakin2018adversarial,dong2018boosting,madry2018towards,xie2019improving} are available in our proposed method(experiments shown in Appendix A). The entire process is shown in Algorithm \ref{ag1}.

%-------------------------------------------------------------------------

\section{Experiments}\label{section5}
In this section, we first describe the implementation details and metrics that we used to generate the FOUND. Next, we present the results of a comprehensive comparison between FOUND and state-of-the-art methods. Then, we systematically conduct ablation experiments and analyze the effectiveness of each proposed method. Finally, we apply FOUND to the black-box deepfake models and analyze the results.

\subsection{Implementation Details and Metrics}
We use CelebA\cite{liu2015deep} as the only dataset to generate FOUND, which contains a total of over 200,000 facial images. We use the first 128 images from the training set as the training images and evaluate the disruption performance in various metrics on all facial images and all possible target attributes from the CelebA testset, the LFW dataset\cite{huang2008labeled}, and the facial images dataset captured from FF++ original videos(FF++O)\cite{rossler2019faceforensics++}, respectively. We use StarGAN\cite{choi2018stargan}, AGGAN\cite{tang2019attention}, AttGAN\cite{he2019attgan}, and HiSD\cite{li2021image} as the ensemble deepfake models to generate FOUND. We evaluate FOUND on these models, and further use StarGAN-v2\cite{choi2020stargan} to evaluate FOUND's performance on black-box deepfake models. StarGAN-v2 is trained on the CelebA-HQ dataset, and the other models are trained on the CelebA dataset. Notice that once we have finished training FOUND, we do not use any images for further training.

\begin{table*}[htpb]
\caption{Quantitative comparisons between CMUA and proposed FOUND. $Avg$ means the average results of each metric, the same hereinafter. Best results are emphasized in \textbf{bold}.}

\label{table1}
\centering
\resizebox{\hsize}{!}{
\begin{tabular}{c|c|cc|cc|cc|cc|cc|cc|cc}
\hline
\multicolumn{1}{c}{\multirow{2}{*}{\textbf{Dataset}}} & \multicolumn{1}{c}{\multirow{2}{*}{\textbf{Model}}} & \multicolumn{2}{c}{\textbf{$L_{2mask}\uparrow$}} & \multicolumn{2}{c}{\textbf{$SR_{mask}\uparrow$}} & \multicolumn{2}{c}{\textbf{FID$\uparrow$}} & \multicolumn{2}{c}{\textbf{PSNR$\downarrow$}} & \multicolumn{2}{c}{\textbf{SSIM$\downarrow$}}  & \multicolumn{2}{c}{\textbf{ACS$\downarrow$}} & \multicolumn{2}{c}{\textbf{TFHC}}  \\
\cline{3-16}
\multicolumn{1}{c}{} & \multicolumn{1}{c}{} & CMUA & \multicolumn{1}{c}{FOUND}   & CMUA & \multicolumn{1}{c}{FOUND} & CMUA & \multicolumn{1}{c}{FOUND} & CMUA & \multicolumn{1}{c}{FOUND} & CMUA & \multicolumn{1}{c}{FOUND} & CMUA  & \multicolumn{1}{c}{FOUND} & CMUA & FOUND \\ 
\hline
\multirow{5}{*}{CelebA}
& StarGAN & 0.21 & \textbf{0.24} & 100.00\% & 100.00\% & 329.09 & \textbf{423.96}& 13.22 & \textbf{12.03} & 0.59 & \textbf{0.53}  & 0.366 & \textbf{0.001} & 67.2\%$\rightarrow$14.4\% & \textbf{67.2\%$\rightarrow$0.00\%} \\
& AGGAN & \textbf{0.25} & 0.23 & 99.91\% & \textbf{100.00\%} & 167.98  & \textbf{297.06}& 15.95 &\textbf{ 15.83} & 0.61 & \textbf{0.58}  & 0.745 & \textbf{0.541} &  67.8\%$\rightarrow$44.6\%&\textbf{ 67.8\%$\rightarrow$19.7\%} \\
& AttGAN & 0.10 & \textbf{0.18} & 86.49\% & \textbf{95.12\%} & 180.05  & \textbf{207.09}& 19.53 & \textbf{16.70} & 0.76 & \textbf{0.71}  & 0.673 & \textbf{0.605} & \textbf{ 61.0\%$\rightarrow$26.4\%} & 61.0\%$\rightarrow$33.9\% \\
& HiSD & \textbf{0.24} & 0.15 & \textbf{99.72\%}  & 99.09\% & \textbf{177.54} & 165.71& \textbf{15.72} & 16.94 & \textbf{0.70} & 0.76  & \textbf{0.295} & 0.774 &  \textbf{65.2\%$\rightarrow$10.1\%} & 65.2\%$\rightarrow$38.0\% \\
& Avg & 0.20 & 0.20 & 96.53\%  & \textbf{98.55\%}  & 213.67 & \textbf{273.46} & 16.11 & \textbf{15.38} & 0.67 & \textbf{0.65} & 0.520 & \textbf{0.480} & $\downarrow$41.43\% &\textbf{ $\downarrow$42.40\%} \\ 
\hline
\multirow{5}{*}{LFW} 
& StarGAN & 0.21 & \textbf{0.24} & 100.00\% & 100.00\% & 274.34 & \textbf{357.82} & 13.01 & \textbf{12.66} & 0.58 &\textbf{ 0.54}  & 0.246 & \textbf{0.002} & 44.0\%$\rightarrow$13.5\% & \textbf{44.0\%$\rightarrow$0.1\%} \\
& AGGAN & 0.25 & 0.25 & 99.48\% & \textbf{100.00\%} & 155.64 & \textbf{226.97}& 15.97 & \textbf{15.52} & 0.61 & \textbf{0.58}  & 0.703 & \textbf{0.507} &  52.7\%$\rightarrow$38.5\%  & \textbf{52.7\%$\rightarrow$15.2\%} \\
& AttGAN & 0.12 &\textbf{ 0.18} & 95.14\% & \textbf{99.87\%} & 197.37 & \textbf{248.56} & 18.79 & \textbf{16.50} & 0.74 & \textbf{0.70} & 0.552 & \textbf{0.234} &  34.2\%$\rightarrow$20.3\%  & \textbf{34.2\%$\rightarrow$13.0\%}\\
& HiSD & \textbf{0.12} & 0.10 & \textbf{97.42\%} & 91.98\% & \textbf{202.21} & 137.55 & \textbf{16.28} & 17.61 & \textbf{0.70} & 0.78 & \textbf{0.495} & 0.810 &  \textbf{50.2\%$\rightarrow$32.6\%}  & 50.2\%$\rightarrow$44.6\% \\
& Avg & 0.18 & \textbf{0.19} & \textbf{98.01\%} & 97.96\% & 207.39 & \textbf{242.72} & 16.01 & \textbf{15.57} & 0.66 & \textbf{0.65} & 0.499 & \textbf{0.388} &  $\downarrow$19.05\% & \textbf{ $\downarrow$27.1\%} \\
\hline
\multirow{5}{*}{FF++O} 
& StarGAN & 0.21 & \textbf{0.23} & 100.00\% & 100.00\% & 336.31 & \textbf{409.45}& 13.48 & \textbf{12.92} & 0.60 & \textbf{0.55}  & 0.277 & \textbf{0.010} &  56.2\%$\rightarrow$12.1\% & \textbf{56.2\%$\rightarrow$0.0\%} \\
& AGGAN & \textbf{0.25} & 0.23 & 100.00\% & 100.00\% & 214.54 & \textbf{375.07}& \textbf{15.51} & 15.78 & 0.60 & \textbf{0.56}  & 0.504 & \textbf{0.288} & 45.3\%$\rightarrow$15.0\% & \textbf{45.3\%$\rightarrow$3.1\%} \\
& AttGAN & 0.14 & \textbf{0.20} & 90.25\% & \textbf{98.49\%} & 207.10 & \textbf{260.82}& 18.36 & \textbf{15.81} & 0.73 & \textbf{0.68}  & 0.527 & \textbf{0.455} & \textbf{30.3\%$\rightarrow$12.0\%} & 30.3\%$\rightarrow$14.8\% \\
& HiSD & \textbf{0.14} & 0.10 & \textbf{99.45\%} & 91.79\% & \textbf{173.99} & 149.66 & \textbf{16.17} & 18.04 & \textbf{0.73} & 0.81  & \textbf{0.423} & 0.750 & \textbf{41.7\%$\rightarrow$20.2\%} & 41.7\%$\rightarrow$25.5\% \\
& Avg & 0.19 & 0.19 & 97.43\% & \textbf{97.57\%} & 232.99 & \textbf{298.75} & 15.88 & \textbf{15.64} & 0.67 & \textbf{0.65} & 0.433 & \textbf{0.376} &  $\downarrow$28.55\%  &  \textbf{ $\downarrow$32.5\%}   \\
\hline
\end{tabular}}
\end{table*}

We set the perturbation upper bound $\epsilon$ to 0.05. During the entire attacking process, the maximum iteration number $T$ of the End-to-End attack is set to 50, and the iteration number $K$ of the feature-ensemble attack is set to 5. The hyperparameter $\lambda$ and learning rate $\alpha$ in Section \ref{sec4.1} are set to 0.01 and 0.001, respectively. The batch size of the facial images is set to 8.

In CMUA-Watermark\cite{huang2022cmua}, authors consider it inappropriate to simply use $L_{2}$ loss to measure the generation quality before and after disruption, the differences between the original image and the original fake image need to be measured. They propose a $L_{2mask}$ loss metric to measure the disruption, when $L_{2mask}$ is larger than 0.05, it is considered a successful deepfake disruption. We follow using the $L_{2mask}$ loss and calculate the success rate $SR_{mask}$. Then, we use PSNR and SSIM that are commonly used in image similarity measurement to evaluate the similarity between images generated before and after the disruption. The lower the score, the lower the image similarity. We further use FID\cite{heusel2017gans} to measure the quality of the fake image after disruption, which uses the Inception-v3\cite{szegedy2016rethinking} network's intermediate features to measure the similarity between the images. The higher the FID value, the lower the similarity. Finally, we use HyperFAS\footnote{\url{https://github.com/zeusees/HyperFAS}}, an open-source liveness detection system, to measure the quality. We measure the \textbf{A}verage \textbf{C}onfidence \textbf{S}core(ACS) of liveness detection and the decline degree of the \textbf{T}rue \textbf{F}ace with \textbf{H}igh \textbf{C}onfidence(TFHC) before and after the disruption. The lower the ACS or the more decline of the TFHC, the better the disruption effect. All experiments were performed under the same settings.

\begin{table}[htpb]
    \caption{Run-time comparison between CMUA and FOUND.}
    \centering
    \setlength{\tabcolsep}{5mm}{
        \label{table2}
        \begin{tabular}{c|cc} 
            \hline
            \textbf{Method} & CMUA & FOUND \\ 
            \hline
            \textbf{Time}   & $>7h$   & $0.71h$  \\ 
            \hline
        \end{tabular}
    }
\end{table}

\subsection{Comparison with State-Of-The-Art Universal Disruptor}
We conduct massive experiments to demonstrate the effectiveness of our proposed FOUND under various metrics in different datasets and models. We show the quantitative results of the comparison of FOUND and state-of-the-art CMUA-Watermark\cite{huang2022cmua} in Table \ref{table1}.

\begin{figure}[htpb]
    \centering
\epstopdfsetup{outdir=./}
    \includegraphics[width=1.0\columnwidth]{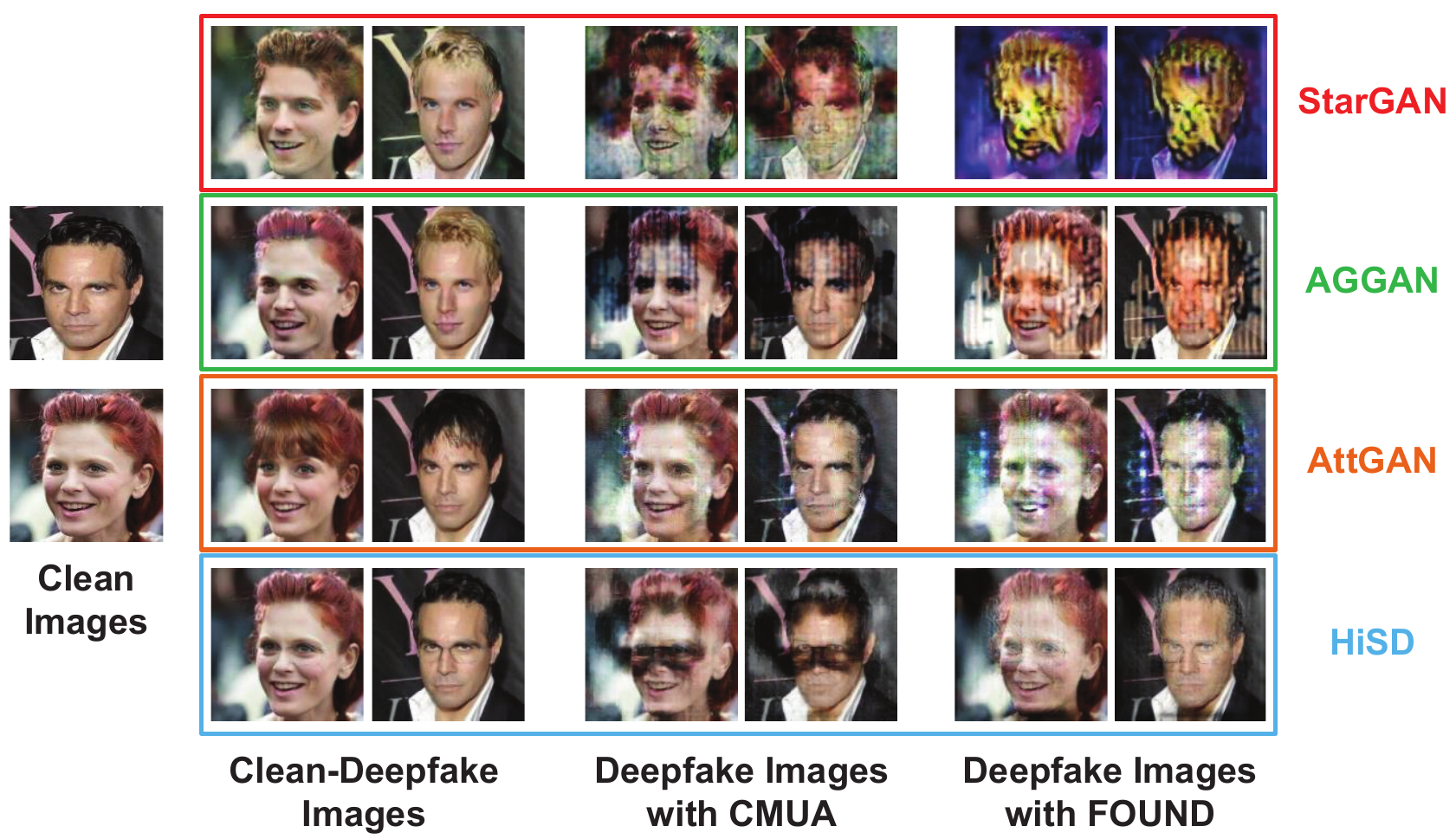}
    \caption{Illustration of examples with CMUA and FOUND.}
\label{fig_show_results}
\end{figure}

The proposed FOUND performs well on CelebA, LFW, and FF++O datasets. Except for slightly underperforming CMUA on HiSD, FOUND outperforms CMUA on StarGAN, AGGAN, and AttGAN, and achieves a better average disruption success rate. This shows that our proposed FOUND has a more balanced effect on the ensemble models than CMUA. It is worth mentioning that FOUND's effect is significantly higher than CMUA's on most FID metrics, indicating that the image quality after FOUND disruption is worse than that of CMUA though they have a similar success rate. When facing the liveness detection model HyperFAS, the confidence degree of the deepfake face disrupted by FOUND decreases more than that of CMUA, which shows that the disrupted deepfake face by FOUND is harder to distinguish than that of CMUA. In Figure \ref{fig_show_results}, we show the disrupted deepfake images by CMUA and FOUND respectively. To sum up, FOUND has realized the cross-attribute, cross-image, cross-dataset, and cross-model universal disruption of DeepFakes, and is further improved compared with CMUA.

On the other hand, benefit from the parallel ensemble attack, generating FOUND is significantly faster than CMUA. CMUA adopts a sequential generation framework, which attacks four models successively, and superimposes the perturbation of each model according to a certain proportion. In order to get the best proportion, they use the NNI tool provided by Microsoft\footnote{\url{https://github.com/microsoft/nni}} and run the program several times with the TPE algorithm\cite{bergstra2011algorithms}. Table \ref{table2} shows the efficiency of generating a universal disruptor between FOUND and CMUA. We follow the settings and train CMUA on three NVIDIA RTX 3090 GPUs and carry out 1000 iterations to find the proportion. Each task takes 75 seconds on average, so the total processing time exceeds 7 hours. In contrast, FOUND generation only takes about 0.71 hours(2558 seconds) in total using only one GPU, greatly improving the generation efficiency of the universal disruptor.

\subsection{Ablation Study}\label{section5.3}

In this section, we evaluate and analyze the effectiveness of each proposed module on CelebA test set using $L_{2mask}$, $SR_{mask}$ and FID.

We first analyze the effectiveness of End-to-End ensemble strategies. As shown in Table \ref{table3}, both Output-Ensemble and Loss-Ensemble cannot successfully disrupt AttGAN, but the success rate of disrupting StarGAN, AGGAN and HiSD is high and obtain a very large $L_{2mask}$ loss. This is because the other three models are more vulnerable to attack than AttGAN. If we simply aggregate functions and calculate the gradient of the entire tasks, the gradient direction will slide to the sub-task that is easier to converge, which is consistent with the discussion in Section \ref{sec4.2}. Using Gradient-Ensemble significantly improves the attack effect on AttGAN, and balances the $L_{2mask}$ value of the other three to a certain extent, which shows that the ensemble of gradients of sub-tasks can effectively avoid falling into local optimum, thus improving the overall disruption effect on the ensemble models.

\begin{table}[t]
    \caption{Comparisons of End-to-End(EE) Logit-Ensemble(LE), Output-Ensemble(OE), and Gradient-Ensemble(GE) strategies. Best results are emphasized in \textbf{bold}.}
    \label{table3}
    \resizebox{\columnwidth}{!}{
        \centering
        \begin{tabular}{c|c|c|c|c|c|c} 
        \hline
        \multicolumn{1}{c}{\multirow{2}{*}{\textbf{Method}}} & \multicolumn{1}{c}{\multirow{2}{*}{\textbf{Metric}}} & \multicolumn{5}{c}{\textbf{Model}}\\
        \cline{3-7}
        \multicolumn{1}{c}{} & \multicolumn{1}{c}{} & StarGAN & AGGAN & AttGAN & HiSD & Avg  \\ 
        \hline
        \multirow{3}{*}{EE(LE)}
        & $SR_{mask}\uparrow$ & 100.00\% & 99.98\% & 0.01\% & 99.79\% & 74.95\% \\
        & $L_{2mask}\uparrow$    & \textbf{1.04} & 0.69 & 0.00 & 0.31 & \textbf{0.51} \\
        & FID$\uparrow$    & \textbf{423.00} & 112.53 & 17.42 & 200.40 & 188.34 \\ 
        \hline
        \multirow{3}{*}{EE(OE)}
        & $SR_{mask}\uparrow$ & 100.00\% & \textbf{99.99\%} & 0.66\% & \textbf{99.93\%} & 75.15\% \\
        & $L_{2mask}\uparrow$ & 1.03 & 0.63 & 0.01 & 0.27 & 0.49 \\
        & FID$\uparrow$    & 418.09 & \textbf{122.91} & 22.94 & 230.95 & 198.72 \\ 
        \hline
        \multirow{3}{*}{EE(GE)}
        & $SR_{mask}\uparrow$ & 100.00\% & 99.97\% & \textbf{66.91\%} & 99.90\% & \textbf{91.70\%} \\
        & $L_{2mask}\uparrow$ & 0.88 & \textbf{0.70} & \textbf{0.09} & \textbf{0.32} & 0.50 \\
        & FID$\uparrow$    & 379.59 & 112.24 & \textbf{121.56} & \textbf{244.99} &     \textbf{214.60} \\ 
        \hline
        \end{tabular}
    }
\end{table}

Then, we analyze the effect of Feature-Ensemble, and the results are shown in Table \ref{table4}. Compared with the End-to-End attacks in Table \ref{table3}, the ensemble attack on the feature extractors of the four models achieves a significantly higher disruption success rate than the various End-to-End disruptions, which shows that disrupting feature extractors is actually the key to disrupting deepfake models. Figure \ref{fig2.2} also indicates this opinion. 
% From the perspective of model complexity, the feature extractor is a part of the deepfake model with a smaller structure and parameter difference than those of the entire deepfake model. Therefore, the derivation of the ensemble feature extractors is simpler than that of the whole process, and it is easier to get the optimization results. 
Furthermore, we combine the Feature-Ensemble with End-to-End Ensemble and find that the disruption success rates of all ensemble strategies have been significantly improved. It shows that disrupting the complete deepfake process on the basis of disrupting feature extractors can significantly alleviate the structural and functional conflicts between models, which further reflects the importance of feature extractors in the deepfake process. The combination of Feature-Ensemble and End-to-End Gradient-Ensemble achieves the best disruption effect, so each method module we propose is effective.

% \subsubsection{Performance of Various Adversarial Attacks}
% \begin{table}
% \resizebox{\columnwidth}{!}{
% \centering
% \begin{tabular}{c|ccccc} 
% \hline
% \multicolumn{1}{c}{\multirow{2}{*}{Method}} & \multicolumn{5}{c}{Models}             \\
% \multicolumn{1}{c}{}                        & StarGAN & AGGAN & AttGAN & HiSD & Avg  \\ 
% \hline
% BIM & 100.00\% & 100.00\% & 91.92\% & 98.13\% & 97.51\% \\
% PGD & 100.00\% & 100.00\% & 92.95\% & 98.42\% & 97.84\% \\
% MI  & 100.00\% & 100.00\% & 91.57\% & 98.32\% & 97.47\% \\
% DI  & 100.00\% & 100.00\% & 93.74\% & 98.23\% & 97.99\% \\
% MIDI & \textbf{100.00\%} & \textbf{100.00\%} & \textbf{94.70\%} & \textbf{99.09\%} & \textbf{98.45\%} \\
% \hline
% \end{tabular}}
% \caption{Results.   Ours is better.}
% \end{table}

\subsection{Disrupting Black-Box DeepFakes}

\begin{figure}[t]
    \centering
\epstopdfsetup{outdir=./}
    \includegraphics[width=0.7\columnwidth]{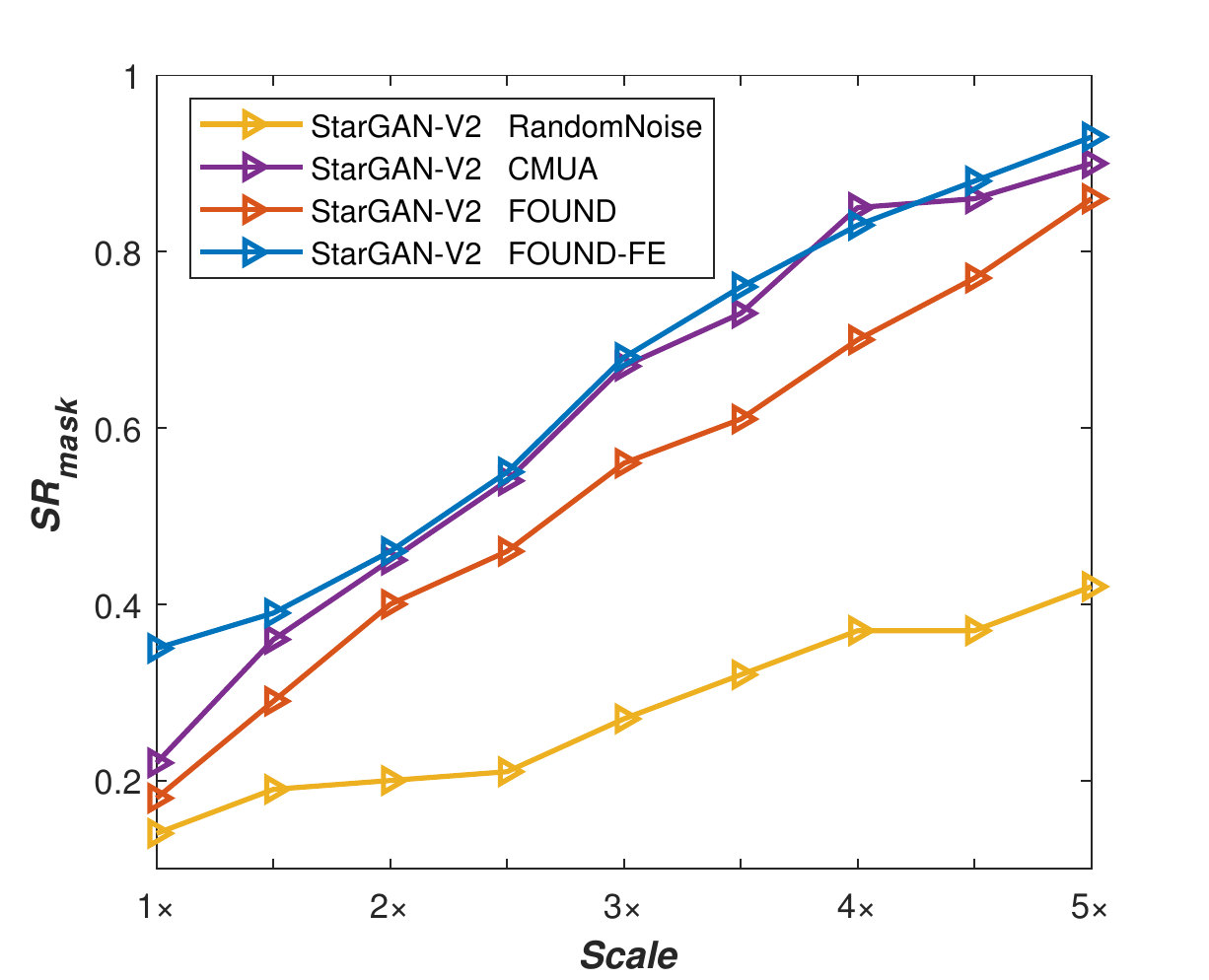}
    \caption{Results of different disruptors on StarGAN-v2.}
\label{fig5}
\end{figure}

In this section, we evaluate the black-box transferability of CMUA and FOUND on StarGAN-v2. We compare $SR_{mask}$ of random Gaussian noise, CMUA, and FOUND under different perturbation scales. Results are shown in Figure \ref{fig5}. Under the initial perturbation scale, the black-box transferability of all methods decreases significantly, which shows that transferring deepfake adversarial perturbation to black-box models is more difficult than that on classification models. However, results indicate that disrupting feature extractors of the deepfake model can not only improve the white-box universality by further combining End-to-End ensemble attacks but also improve the black-box transferability. FOUND-FE which only disrupts cross-model feature extractors achieves the best effect under almost all perturbation scales. With the gradual amplification of the perturbation scale, the increase in the success rate of FOUND is less than that of CMUA but is faster than random Gaussian noise. This indicates that there is a trade-off between the strong white-box universality and the strong black-box transferability of the universal disruptor. 
% More powerful black-box transfer disruptors need to be explored in our future works. 

\begin{table}[t]
    \caption{Comparisons of Feature-Ensemble(FE) and the combination of the other three End-to-End ensemble strategies. Best results are emphasized in \textbf{bold}.}
    \label{table4}
    \resizebox{\columnwidth}{!}{
        \centering
        \begin{tabular}{c|c|c|c|c|c|c} 
            \hline
            \multicolumn{1}{c}{\multirow{2}{*}{\textbf{Method}}} & \multicolumn{1}{c}{\multirow{2}{*}{\textbf{Metric}}} & \multicolumn{5}{c}{\textbf{Model}}                \\
            \cline{3-7}
            \multicolumn{1}{c}{} & \multicolumn{1}{c}{} & StarGAN  & AGGAN    & AttGAN  & HiSD    & Avg      \\ 
            \hline
            \multirow{3}{*}{FE} 
            & $SR_{mask}\uparrow$ & 99.99\% & 100.00\% & 89.42\% & 93.36\% & 95.69\%  \\
            & $L_{2mask}\uparrow$ & 0.22 & 0.21 & 0.13 & 0.10 & 0.17 \\
            & FID$\uparrow$ & 447.22 & 251.14 & 162.66 & 139.80 & 250.21 \\ 
            \hline
            \multirow{3}{*}{FE+EE(LE)} 
            & $SR_{mask}\uparrow$ & 100.00\% & 100.00\% & 90.11\% & 95.10\% & 96.30\%  \\
            & $L_{2mask}\uparrow$ & \textbf{0.26} & 0.22 & 0.14 & 0.10 & 0.18 \\
            & FID$\uparrow$ &\textbf{ 472.55} & 281.73 & 160.98 & 143.19 & 264.61 \\ 
            \hline
            \multirow{3}{*}{FE+EE(OE)} 
            & $SR_{mask}\uparrow$ & 100.00\% & 100.00\% & 89.38\% & 95.24\% & 96.16\%  \\
            & $L_{2mask}\uparrow$ & 0.24 & 0.22 & 0.14 & 0.10 & 0.18 \\
            & FID$\uparrow$ & 466.84 & 274.65 & 162.73 & 144.18 & 262.10 \\ 
            \hline
            \multirow{3}{*}{FE+EE(GE)(\textbf{FOUND})}                           
            & $SR_{mask}\uparrow$ & \textbf{100.00\%} & 100.00\% & \textbf{95.12\%} & \textbf{99.09\%} & \textbf{98.55\%}  \\
            & $L_{2mask}\uparrow$ & 0.24 & \textbf{0.23} & \textbf{0.18} & \textbf{0.15} & \textbf{0.20} \\
            & FID$\uparrow$ & 423.96 & \textbf{297.06} & \textbf{207.09} & \textbf{165.71} & \textbf{273.46} \\
            \hline
        \end{tabular}
    }
\end{table}

\section{Conclusion and Future Works}
In this paper, we introduce a simple yet effective two-stage universal deepfake disruption method named FOUND. To the best of our knowledge, we are the first to analyze and attack ensemble feature extractors in deepfake disruption. We further propose Gradient-Ensemble in the End-to-End ensemble attack to solve the complex multi-deepfake model optimization problem. Experiments demonstrate that our proposed method realizes a new state-of-the-art universal deepfake disruption. It also shows that disrupting feature extractors can not only improve the white-box universality by further combining End-to-End ensemble disruption but also has the potential in improving the black-box transferability. 
% This provides new insights for follow-up research on deepfake active defenses.

We only focus on ensemble models of the attribute modification category in this work, due to the more complex and different structures of cross-category DeepFakes. Still, the proposed novel disruption framework has the potential to be applied to the broader DeepFakes, because the facial image downsampling process is important and ubiquitous in various DeepFakes. Besides, our idea can provide new insights and baselines for deepfake disruption against black-box models. In future works, we wish to obtain a broader and generalized deepfake universal disruptor, which will thoroughly avoid deepfake privacy risks.

% The limitation of this paper is that we only focus on ensemble models of the attribute modification category due to the more complex and different structures of cross-category DeepFakes. Still, the proposed novel disruption framework has the potential to be applied to the broader DeepFakes, because the facial image downsampling process is important and ubiquitous in various DeepFakes. Besides, our idea can provide new insights and baselines for deepfake disruption against black-box models. In future works, we wish to obtain a broader and generalized deepfake universal disruptor, which will thoroughly avoid deepfake privacy risks.

%%%%%%%%% REFERENCES
{\small
\bibliographystyle{ieee_fullname}
\bibliography{PaperForReview}
}

\end{document}